\documentclass{INTERSPEECH2023}

\interspeechcameraready 

\usepackage{hyperref}
\hypersetup{
    colorlinks=true,
    urlcolor=black,
    }

\title{EMNS /Imz/ Corpus: An emotive single-speaker dataset for narrative storytelling in games, television and graphic novels}
\name{Kari A Noriy$^1$$^2$, Xiaosong Yang$^1$$^2$, Jian Jun Zhang$^1$$^2$}
\address{
  $^1$Centre for Digital Entertainment, Bournemouth University, UK
  $^2$National Centre for Computer Animation, Bournemouth University, UK
  }
\email{knoriy@bournemouth.ac.uk, xyang@bournemouth.ac.uk, jzhang@bournemouth.ac.uk}

\begin{document}

\maketitle

\begin{abstract}

The increasing adoption of text-to-speech technologies has led to a growing demand for natural and emotive voices that adapt to a conversation's context and emotional tone. The Emotive Narrative Storytelling (EMNS) corpus is a unique speech dataset created to enhance conversations' expressiveness and emotive quality in interactive narrative-driven systems. The corpus consists of a 2.3-hour recording featuring a female speaker delivering labelled utterances. It encompasses eight acted emotional states, evenly distributed with a variance of 0.68\%, along with expressiveness levels and natural language descriptions with word emphasis labels. The evaluation of audio samples from different datasets revealed that the EMNS corpus achieved the highest average scores in accurately conveying emotions and demonstrating expressiveness. It outperformed other datasets in conveying shared emotions and achieved comparable levels of genuineness. A classification task confirmed the accurate representation of intended emotions in the corpus, with participants recognising the recordings as genuine and expressive. Additionally, the availability of the dataset collection tool under the Apache 2.0 License simplifies remote speech data collection for researchers.
\end{abstract}

\section{Introduction}

Emotive Narrative Storytelling has emerged as a captivating approach in speech synthesis, aiming to generate more expressive and emotionally engaging conversations for interactive narrative-driven systems. The ability to convey emotions effectively through synthesised speech is crucial for creating engaging user experiences in various applications, including virtual assistants, interactive storytelling, and entertainment platforms.

The emotional expressiveness of speech plays a vital role in human communication, enabling us to convey rich and nuanced meanings beyond the literal interpretation of words. Traditional speech synthesis systems often struggle to capture the intricacies of human emotions, resulting in monotonous and robotic-sounding voices. The development of the EMNS corpus aims to address this limitation by providing a comprehensive and diverse dataset incorporating a range of emotions, linguistic variations,  tonal variance and natural language descriptions.

In this paper, we present the EMNS corpus, a novel speech dataset explicitly designed to enhance synthesised speech's expressiveness and emotional impact. We conducted a series of experiments to evaluate the efficacy of the corpus, involving a diverse group of participants, including both native and non-native English speakers. Through these experiments, we aimed to capture valuable insights into participants' self-reported emotional states, their evaluations of recordings derived from different datasets, and their ability to identify and discern emotions and expressiveness levels within the EMNS corpus.

In section \ref{sec:emns_corpus}, we provide an in-depth description of the EMNS corpus, encompassing its composition, emotional categories, expressiveness levels, and natural language descriptions supplemented with word emphasis labels. Section \ref{sec:experiments}, outlines the experimental design and methodology employed to evaluate the effectiveness of the EMNS corpus and present the outcomes of these experiments, including a comparative analysis of the EMNS corpus against other existing datasets, the accuracy of emotion identification, and participants' perception of expressiveness and genuineness. Finally, we summarise the findings, explore their implications for speech synthesis research, and propose potential avenues for further enhancing the EMNS corpus.

\section{Related work}

Over the years, several influential speech datasets have been introduced \cite{dataset:ljspeech17, dataset:css10, dataset:libriTTS, dataset:mcu-arctic}. These datasets have served as widely used benchmarks for evaluating text-to-speech techniques \cite{paper:adaspeech_2021, paper:diff-tts:_2021, paper:fastspeech_2021, paper:lightspeech:_2021} due to their extensive samples, accurate transcriptions, and high-quality audio recordings. However, these datasets have yet to evolve to meet the changing demands of state-of-the-art models, which require additional paralinguistic features such as emotion and variations in perceived emotional intensity.

To address this gap and improve the modality of speech datasets by providing paralinguistic information, several publicly available datasets have been proposed \cite{dataset:RAVDESS, dataset:GEMEP, dataset:CREMA-D, dataset:MSP-IMPROV, dataset:EmoV-DB}. However, these datasets have certain limitations that make them less suitable for high-end interactive systems. For instance, RAVDESS \cite{dataset:RAVDESS} is a multi-modal emotional speech and song database that offers a large sample size but lacks phonetic diversity, as it includes only two prompts read by all participants. This limited phonetic diversity hinders the ability of speech synthesis systems to generalise and produce meaningful results for unseen prompts. Similarly, the CREMA-D \cite{dataset:CREMA-D} corpus consists of speech from various ethnic backgrounds but also suffers from a lack of phonetic diversity and provides only twelve unique prompts.
GEMEP \cite{dataset:GEMEP} follows a similar approach to RAVDESS and CREMA-D, providing labelled emotions from French speakers at three intensity levels. However, due to its improvised nature, the dataset contains inconsistent recording quality, such as low volume, truncated audio, or stuttering. MSP-IMPROV \cite{dataset:MSP-IMPROV} improves on the aforementioned datasets by providing a multi-modal emotional corpus consisting of audio and video recordings with 652 target prompts. However, this dataset is not suitable for speech synthesis due to the recording setup and the presence of improvised overlapping conversation in the recordings. Similarly, EmoV\_DB \cite{dataset:EmoV-DB}, although recorded in an anechoic chamber, still contains unwanted noises such as knocks and bumps, making it less ideal for high-quality speech synthesis.

In conclusion, while RAVDESS \cite{dataset:RAVDESS}, CREMA-D \cite{dataset:CREMA-D}, GEMEP \cite{dataset:GEMEP}, and MSP-IMPROV \cite{dataset:MSP-IMPROV} provide valuable emotive data, they fall short in areas such as phonetic diversity and noise-free recordings. Among the existing datasets, EmoV\_DB \cite{dataset:EmoV-DB} achieves results closest to a modern dataset for interactive systems, using phonetically balanced sentences. However, it still suffers from unwanted noises in the recording setup. Table \ref{tab:dataset_list} presents the statistics for these publicly available datasets.
Our corpus, EMNS, aims to overcome these limitations and promote dynamic speech synthesis in interactive narrative-driven systems by providing a diverse dataset free of noise, including natural language descriptions and speaker emotion labels.

\begin{table*}[th!]
  \caption{Publicly available speech datasets. Where NLD is Natural Language Description with or without word emphasis and EL is Expressiveness Levels, we consider neutral and unlabelled EL as one expressed level.}
  \label{tab:dataset_list}
  \centering
  \begin{tabular}{ lllclc }
    \toprule
    \textbf{Name} & \textbf{Language} & \textbf{No. Emotions} & \textbf{NLD} & \textbf{EL}\\
    \midrule
    ESD\cite{dataset:ESD}                           & Multilingual      & 5             & No            & 1\\
    RAVDESS\cite{dataset:RAVDESS}                   & English           & 8             & No            & 3\\
    CREMA-D\cite{dataset:CREMA-D}                   & English           & 6             & No            & 3\\
    GEMEP\cite{dataset:GEMEP}                       & French            & 15            & No            & 3\\
    MSP-IMPROV\cite{dataset:MSP-IMPROV}             & English           & 4             & No            & 1\\
    EmoV-DB\cite{dataset:EmoV-DB}                   & English           & 5             & No            & 4\\
    \midrule
    \textbf{EMNS}                                   & \textbf{English}  & \textbf{8}    & \textbf{Yes}  & \textbf{10}\\
    \bottomrule
  \end{tabular}
\end{table*}

\section{EMNS corpus}\label{sec:emns_corpus}

\begin{figure}[t]
  \centering
  \includegraphics[width=0.9\linewidth]{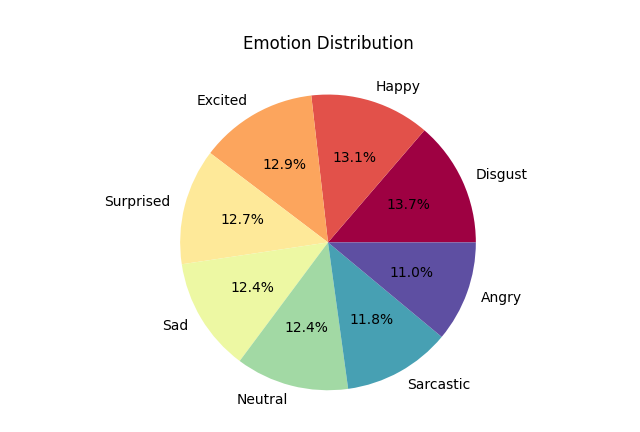}
  \caption{The representation of emotions within the EMNS dataset.}
  \label{fig:emotion_dist}
\end{figure}

The EMNS corpus addresses the limitations of existing datasets and aims to facilitate dynamic speech synthesis in interactive narrative-driven systems. It consists of 2.3 hours of labelled utterances delivered by a female speaker, offering a diverse range of emotional states as depicted in Figure \ref{fig:emotion_dist}. This variety allows for nuanced and diverse expressions, enabling speech generation with varying degrees of emotional impact. The dataset provides ten emotional intensity levels, allowing for precise control over the emotional expression.

In addition to its comprehensive emotional coverage, the EMNS dataset offers valuable features such as natural language descriptions. This aspect enhances the dataset's usefulness in generating speech that not only conveys emotional states but also incorporates contextually appropriate and meaningful language. By including this information, the EMNS dataset facilitates the development of speech synthesis systems that can produce coherent and engaging narratives within interactive applications. Furthermore, the recordings in the EMNS dataset are of high quality and free from unwanted background noises or inconsistencies.

By combining emotional variety, intensity levels, natural word descriptions, and noise-free recordings, the EMNS dataset provides a robust foundation for training and evaluating state-of-the-art text-to-speech techniques. It caters to the evolving demands of modern speech synthesis models, which require paralinguistic features to generate emotionally expressive and engaging speech output. The EMNS dataset opens up new possibilities for creating interactive systems that can dynamically generate speech with nuanced emotions, fostering more immersive and captivating user experiences.

\subsection{Corpus content}

\begin{figure}[t]
  \centering
  \includegraphics[width=0.9\linewidth]{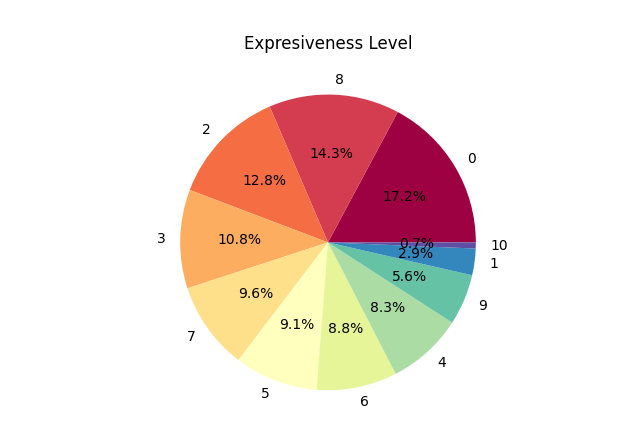}
  \caption{Distribution of expressiveness levels as reported by the voice actors in the EMNS dataset. where 0 represents a neutral state, 5 represents moderate expressiveness, and 10 represents a very expressive state}
  \label{fig:expressiveness_level}
\end{figure}

To ensure a comprehensive range of sentence types, we carefully selected a diverse subset of sentences from Mozilla's Common Voice (CV) dataset. This community-driven dataset consists of phonetically diverse and validated sentences regularly updated with new words. By leveraging this selection approach, the EMNS dataset offers two key advantages: a wide range of phonetic variations and the potential for future expansion.
In the EMNS dataset, emotions are evenly distributed, with a variance of 0.68\%. During the recording process, the voice actor self-reported their expressiveness level on a scale of 0 to 10, where 0 represents a neutral state, and 10 indicates a highly expressive state. This expressive scale of 0 to 10 allows for precise control of the emotional intensity, reducing ambiguity compared to methods such as low, medium, high, and neutral used in other datasets like CREMA-D and GEMEP. Figure \ref{fig:expressiveness_level} illustrates the distribution of expressiveness levels within the dataset. Further details regarding the accuracy of these levels are explored in Section \ref{sec:experiments}.
The natural description provided with each utterance includes the transcription of the recording, the associated emotion, and emphasised words. For example, in the description, \textit{Placing emphasis on Commits, crimes and pennies, \textbf{user\_id} said, "He now commits crimes which center around pennies."} This comprehensive natural description enables a contextually rich understanding of the recorded content, including the specific emotion and emphasised words used.

\subsubsection{Metadata}
The EMNS corpus is distributed in a Tab-Separated Values (TSV) file format along with a WebM subdirectory containing all the recorded utterances. The TSV file includes five essential attributes: Utterance, Description, Emotion, Intensity, and Path to the file, as summarised in Table \ref{tab:metadata}.
The Emotion attribute encompasses eight distinct emotions: happy, sad, angry, excited, sarcastic, neutral, disgusted, and surprised. The selection of these specific emotions is based on careful consideration and is further discussed in Section \ref{section:Choice of Emotions}.
The audio files are stored in the open-source \textit{.webm} format, which is widely used and optimised for web applications. It is supported by modern web browsers such as Firefox, Chrome, and Edge, ensuring convenient accessibility. The corpus releases the audio files as mono-channel, 32-bit audio files at a 48 kHz sample rate, aligning with the specifications of the Common Voice dataset \cite{dataset:common_voice}.

\begin{table}[t]
    \centering
    \caption{The metadata found in the EMNS corpus}
    \begin{tabular}{cp{5cm}}
        \toprule
        \textbf{Title} & \textbf{Description} \\
        \midrule
        Utterance   & A transcription of a given audio file \\
        Description & A description of how the actor sounded and any words they emphasised. \\
        Emotion     & The acted emotion (happy, sad, angry, excited, sarcastic, neutral, disgusted and surprised) \\
        Intensity   & Voice actors self-reported emotion intensity / Expressive Level (EL) \\
        Path        & Path to the audio file \\
        \bottomrule
    \end{tabular}
    \label{tab:metadata}
\end{table}

\subsection{Corpus creating} \label{sec:corpus_Creating}
The EMNS corpus was collected and validated using custom web applications specifically designed to facilitate the remote recording of high-quality audio. To ensure the integrity and reliability of the corpus, a comprehensive validation process was implemented, which involved thorough checks for mispronunciations, unwanted artefacts, and noise. The web platform employed three distinct user roles: Administrator, Actor, and Viewer.
Actors, assuming the role of voice contributors, recorded predefined prompts while being presented with a designated emotion on the screen. The recorded utterances were then subject to review by an administrator responsible for verifying the accuracy of the transcript and identifying any unwanted artefacts. The final dataset ultimately included only recordings that met the predefined quality standards. Utterances that did not meet the established quality criteria were returned to the actors for re-recording, ensuring high audio quality and accuracy in the corpus.
For transparency and reproducibility, the source code of the dataset collection tool (DCT) used in this process is openly available at the following GitHub repository: \anon{\url{https://github.com/knoriy/EMNS-DCT}}.

\subsubsection{Recording setup}
The recording environment was optimised for sound quality with a sound-dampened room and sound-absorbing materials. To capture clear audio, a Shure MV7 USB Microphone with a cardioid pattern was used to help reduce reflected sounds.
The microphone was mounted on a desk tripod with a wind filter. Through experimentation, it was determined that recording audio from a distance of approximately 10 centimetres from the mount produced the best quality audio. Additionally, the microphone was angled at approximately 30 degrees to reduce plosives caused by emotions and words requiring loud, sharp air bursts.

\subsubsection{Choice of emotions} \label{section:Choice of Emotions}
The study of emotion has provided insights into the existence of universally recognized basic emotions across cultures, as classified by Ekman et al. \cite{book:ekman_1999} and Shaver et al. \cite{paper:shaver_emotion_1987}. In line with this understanding, our dataset incorporates sarcasm as an emotion based on valuable industry feedback. Including sarcasm in our list of emotions is crucial due to its unique nature and the complexity it adds to emotional expression. Sarcasm can involve a blend of multiple emotions and is highly context-dependent, making it an essential addition to capturing the intricacies of human emotional experiences.

By including this emotion, we offer researchers and developers a valuable resource to explore and study the interplay between sarcasm and other emotional states. Sarcasm adds a layer of nuance and subtlety to emotional expression, providing an opportunity to enhance the authenticity and realism of interactive systems that generate speech.
The emotions we label in our dataset include Happy, Sad, Angry, Excited, Neutral, Disgusted, Surprised, and Sarcastic. This comprehensive range ensures that our dataset captures a broad range of human emotions, allowing for more nuanced and contextually appropriate speech synthesis. In addition, by incorporating sarcasm into this collection, we enable the development of speech synthesis models that can accurately convey the intricate emotional nuances present in everyday communication, resulting in more engaging and realistic interactions between humans and interactive systems.

\subsubsection{Corpus preprocessing}
The extraction of phoneme duration in the EMNS corpus is carried out using the Montreal Forced Aligner (MFA) \cite{paper:MFA}, a state-of-the-art tool for aligning audio and text to determine time stamps of phonemes in the audio. This process aids speech synthesis research, as accurate measurement of phoneme duration leads to more natural and expressive synthetic speech \cite{paper:fastspeech_2021}.

MFA also identifies silences in the recording, which we used to remove pauses at the beginning and end of the recording and the recorded clicking sound of the mouse used to start and stop the recording. The modified alignments in the EMNS corpus provide researchers with valuable information for enhancing speech synthesis models.

\section{Experiments} \label{sec:experiments}

To evaluate the effectiveness of the EMNS corpus, a series of experiments were conducted involving 19 participants, consisting of both native and non-native English speakers. The objective of the survey was to gather valuable insights into the participants' self-reported emotional states, their evaluations of recordings derived from different datasets, and their ability to recognize and distinguish emotions and expressiveness levels within the EMNS corpus. Notably, the participant pool was carefully balanced, with 32\% being native English speakers and the remaining 68\% representing individuals from diverse countries such as France, Greece, Romania, and Italy. This deliberate inclusion of non-native speakers allowed for a comprehensive assessment of the EMNS corpus's effectiveness across various linguistic backgrounds and cultural influences.
The study findings indicated no significant disparity in emotion perception between the native and non-native English speakers who participated.

\subsection{Emotional states}
To ensure a thorough examination of potential biases in participants' emotion perception during the listening test, they were asked to self-report their short and medium-term emotional states prior to the evaluation. Our findings indicate that 58\% of participants reported feeling content at the time of the test, while 21\% reported being happy. The remaining emotions, such as excited, restless, sad, and worried, were evenly distributed among the participants.
When considering the participants' daily emotional state, 53\% reported feeling content, and 32\% reported being happy. The remaining emotions, including restless and worried, were distributed among the participants. Over the course of a week, the reported emotional states encompassed happiness (24\%), contentment (29\%), excitement (10\%), restlessness (19\%), and worry (19\%). Notably, no significant associations were observed between participants' short and medium-term emotional states that could potentially impact their ability to perceive emotions accurately.

By collecting this detailed emotional data, we aimed to identify and address any potential confounding factors that could influence participants' perception of emotions in the recorded samples.

\subsection{Dataset comparison}
In the study, participants evaluated audio samples from Crema-D, ESD, RAVDESS, MSP-IMPROV, and EMNS to accurately assess their ability to convey shared emotional states. Randomly selected samples from each dataset were rated based on their portrayal of sadness, neutrality, happiness, and anger across all five datasets. Ratings were provided in three categories: conveying shared emotion, expressiveness, and genuineness.
The results indicated that the EMNS corpus achieved the highest average scores for accurately conveying emotions, receiving 41\% of the votes, followed by MSP-IMPROV with 30\%. In terms of expressiveness, EMNS outperformed other datasets with 46\% of the votes, while MSP-IMPROV received 29\%. Regarding genuineness, Crema-D obtained the highest average score at 38\%, with EMNS and MSP-IMPROV following at 22\% and 24\%, respectively.

\subsection{Evaluating EMNS accuracy}
The participants assessed the precision of the recorded emotional state and expressiveness through a classification task. They were required to identify the emotion from the eight options provided in section \ref{section:Choice of Emotions} and rate the expressiveness on a scale of 0 to 10.

\begin{figure}[t]
  \centering
  \includegraphics[width=0.9\linewidth]{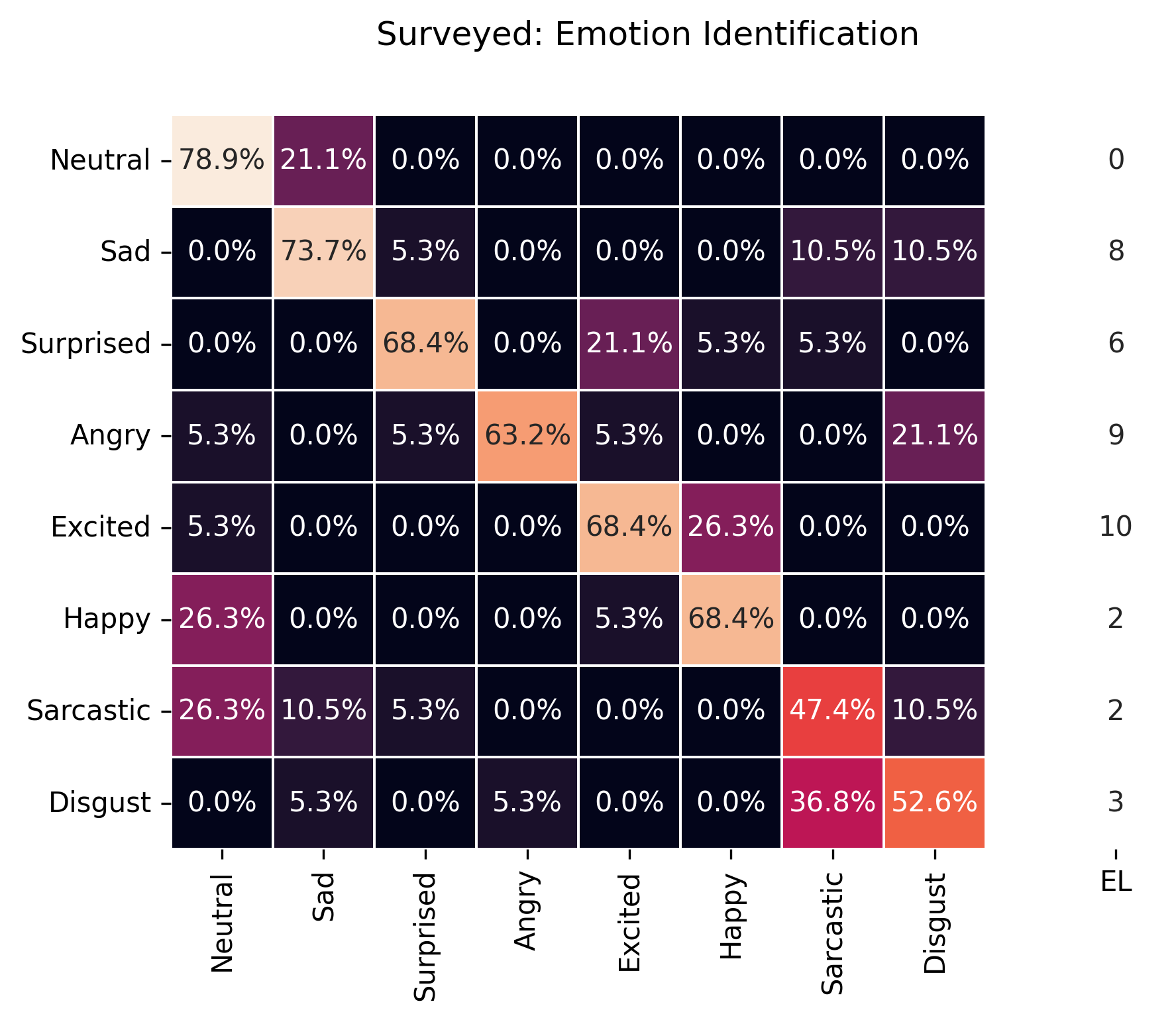}
  \caption{Survey conducted of 19 people to classify the eight emotions from the EMNS Corpus. Where EL is the self-reported Expressiveness Level by the voice actor.}
  \label{fig:surveyed-emotion-classification}
\end{figure}

\begin{figure}[t]
  \centering
  \includegraphics[width=0.9\linewidth]{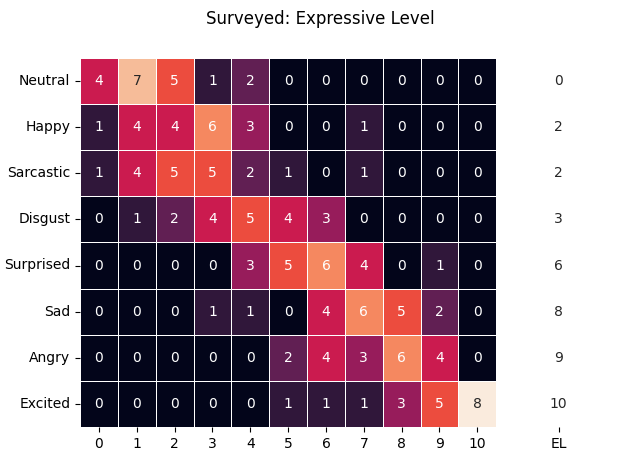}
  \caption{The observed Expressiveness Level (EL) of randomly selected recordings from each emotion.}
  \label{fig:surveyed-expressive-level}
\end{figure}

As illustrated in figures \ref{fig:surveyed-emotion-classification} and \ref{fig:surveyed-expressive-level}. the results of these experiments confirm that the EMNS corpus accurately represents the intended emotions and that listeners perceive the recordings as genuine and expressive

Moreover, figure \ref{fig:surveyed-emotion-classification} shows that participants could accurately identify the acted emotion. The findings also revealed some noteworthy observations. For instance, participants commonly mistook disgust for anger, particularly at low expressiveness levels. Similarly, happy and neutral emotions were often mistaken for each other. Furthermore, we noted that sarcasm was frequently misclassified with other emotions, particularly on low expressiveness levels, likely due to the nature of sarcasm and its contextual nature. Figure \ref{fig:surveyed-expressive-level} shows the recorded samples accurately reflect the perceived expressiveness level (EL) compared to the ground truth.

The successful emotion identification task results demonstrated the EMNS corpus's efficacy as a valuable tool for speech synthesis research. The survey also provided valuable feedback on the accuracy of the emotional representation in the corpus and the perceived expressiveness of the recordings, highlighting areas for potential dataset improvements.


\section{Conclusions}
In this study, we introduced the Emotive Narrative Storytelling (EMNS) corpus, a new speech dataset designed to generate more expressive and emotive conversations for interactive narrative-driven systems. The corpus consists of a 2.3-hour recording of labelled utterances spoken by a female speaker, featuring eight acted emotional states and equal distribution of emotions with a variance of 0.68\%, expressiveness level, and natural language description with word emphasis labels. 

The results of the experiments conducted to evaluate the perceived emotional state of the recordings indicate that the corpus accurately represents the intended emotions and that listeners perceive the recordings as expressive and genuine. Furthermore, the emotion identification task results demonstrate the validity of the EMNS corpus as a valuable tool for speech synthesis research. Additionally, the survey allowed us to assess the accuracy of the emotional representation in the EMNS corpus and the perceived expressiveness of the recordings, providing valuable feedback for further improvements to the dataset.

We make the source code of the dataset collection tool available under the Apache 2.0 Licence \cite{license:apache_2004}, enabling researchers to collect speech data remotely easily.

Overall, the EMNS corpus is a valuable resource for researchers and practitioners in speech synthesis, providing a wide range of phonetically diverse sentences, natural conversational description audio/text pairs, and an equal distribution of emotions. Future work includes expanding the number of samples, adding paralinguistic features as speech synthesis research progresses, and expanding the remote dataset tool to simplify speech data collection further.

\section{Acknowledgements}
This research was supported by The Engineering and Physical Sciences Research Council (EPSRC) by means of a grant awarded to \anon{Kari A Noriy. CDE2: EP/L016540/1 (for 2014/15 cohort and subsequent intakes)}

\bibliographystyle{IEEEtran}
\bibliography{mybib}

\begin{thebibliography}{10}
\providecommand{\url}[1]{#1}
\csname url@samestyle\endcsname
\providecommand{\newblock}{\relax}
\providecommand{\bibinfo}[2]{#2}
\providecommand{\BIBentrySTDinterwordspacing}{\spaceskip=0pt\relax}
\providecommand{\BIBentryALTinterwordstretchfactor}{4}
\providecommand{\BIBentryALTinterwordspacing}{\spaceskip=\fontdimen2\font plus
\BIBentryALTinterwordstretchfactor\fontdimen3\font minus
  \fontdimen4\font\relax}
\providecommand{\BIBforeignlanguage}[2]{{%
\expandafter\ifx\csname l@#1\endcsname\relax
\typeout{** WARNING: IEEEtran.bst: No hyphenation pattern has been}%
\typeout{** loaded for the language `#1'. Using the pattern for}%
\typeout{** the default language instead.}%
\else
\language=\csname l@#1\endcsname
\fi
#2}}
\providecommand{\BIBdecl}{\relax}
\BIBdecl

\bibitem{dataset:ljspeech17}
K.~Ito and L.~Johnson, ``The lj speech dataset,''
  \url{https://keithito.com/LJ-Speech-Dataset/}, 2017.

\bibitem{dataset:css10}
\BIBentryALTinterwordspacing
K.~Park and T.~Mulc, ``{CSS10:} {A} collection of single speaker speech
  datasets for 10 languages,'' \emph{CoRR}, vol. abs/1903.11269, 2019.
  [Online]. Available: \url{http://arxiv.org/abs/1903.11269}
\BIBentrySTDinterwordspacing

\bibitem{dataset:libriTTS}
H.~Zen, V.~Dang, R.~Clark, Y.~Zhang, R.~J. Weiss, Y.~Jia, Z.~Chen, and Y.~Wu,
  ``Libritts: A corpus derived from librispeech for text-to-speech,''
  \emph{Interspeech 2019}, 2019.

\bibitem{dataset:mcu-arctic}
\BIBentryALTinterwordspacing
J.~Kominek and A.~W. Black, ``The {CMU} arctic speech databases,'' in
  \emph{Fifth {ISCA} {ITRW} on Speech Synthesis, Pittsburgh, PA, USA, June
  14-16, 2004}, A.~W. Black and K.~A. Lenzo, Eds.\hskip 1em plus 0.5em minus
  0.4em\relax {ISCA}, 2004, pp. 223--224. [Online]. Available:
  \url{http://www.isca-speech.org/archive\_open/ssw5/ssw5\_223.html}
\BIBentrySTDinterwordspacing

\bibitem{paper:adaspeech_2021}
\BIBentryALTinterwordspacing
Y.~Yan, X.~Tan, B.~Li, T.~Qin, S.~Zhao, Y.~Shen, and T.-Y. Liu, ``Adaspeech 2:
  {Adaptive} {Text} to {Speech} with {Untranscribed} {Data},'' in
  \emph{{ICASSP} 2021 - 2021 {IEEE} {International} {Conference} on
  {Acoustics}, {Speech} and {Signal} {Processing} ({ICASSP})}.\hskip 1em plus
  0.5em minus 0.4em\relax Toronto, ON, Canada: IEEE, Jun. 2021, pp. 6613--6617.
  [Online]. Available: \url{https://ieeexplore.ieee.org/document/9414872/}
\BIBentrySTDinterwordspacing

\bibitem{paper:diff-tts:_2021}
\BIBentryALTinterwordspacing
M.~Jeong, H.~Kim, S.~J. Cheon, B.~J. Choi, and N.~S. Kim, ``Diff-{TTS}: {A}
  {Denoising} {Diffusion} {Model} for {Text}-to-{Speech},'' arXiv, Tech. Rep.
  arXiv:2104.01409, Apr. 2021, arXiv:2104.01409 [cs, eess] type: article.
  [Online]. Available: \url{http://arxiv.org/abs/2104.01409}
\BIBentrySTDinterwordspacing

\bibitem{paper:fastspeech_2021}
\BIBentryALTinterwordspacing
Y.~Ren, C.~Hu, X.~Tan, T.~Qin, S.~Zhao, Z.~Zhao, and T.-Y. Liu, ``{FastSpeech}
  2: {Fast} and {High}-{Quality} {End}-to-{End} {Text} to {Speech},'' arXiv,
  Tech. Rep. arXiv:2006.04558, Mar. 2021, arXiv:2006.04558 [cs, eess] type:
  article. [Online]. Available: \url{http://arxiv.org/abs/2006.04558}
\BIBentrySTDinterwordspacing

\bibitem{paper:lightspeech:_2021}
\BIBentryALTinterwordspacing
R.~Luo, X.~Tan, R.~Wang, T.~Qin, J.~Li, S.~Zhao, E.~Chen, and T.-Y. Liu,
  ``{LightSpeech}: {Lightweight} and {Fast} {Text} to {Speech} with {Neural}
  {Architecture} {Search},'' arXiv, Tech. Rep. arXiv:2102.04040, Feb. 2021,
  arXiv:2102.04040 [cs, eess] type: article. [Online]. Available:
  \url{http://arxiv.org/abs/2102.04040}
\BIBentrySTDinterwordspacing

\bibitem{dataset:RAVDESS}
\BIBentryALTinterwordspacing
S.~R. Livingstone and F.~A. Russo, ``The ryerson audio-visual database of
  emotional speech and song ({RAVDESS}): A dynamic, multimodal set of facial
  and vocal expressions in north american english,'' \emph{{PLOS} {ONE}},
  vol.~13, no.~5, p. e0196391, May 2018. [Online]. Available:
  \url{https://doi.org/10.1371/journal.pone.0196391}
\BIBentrySTDinterwordspacing

\bibitem{dataset:GEMEP}
\BIBentryALTinterwordspacing
T.~B\"{a}nziger, M.~Mortillaro, and K.~R. Scherer, ``Introducing the geneva
  multimodal expression corpus for experimental research on emotion
  perception.'' \emph{Emotion}, vol.~12, no.~5, pp. 1161--1179, 2012. [Online].
  Available: \url{https://doi.org/10.1037/a0025827}
\BIBentrySTDinterwordspacing

\bibitem{dataset:CREMA-D}
H.~Cao, D.~G. Cooper, M.~K. Keutmann, R.~C. Gur, A.~Nenkova, and R.~Verma,
  ``Crema-d: Crowd-sourced emotional multimodal actors dataset,'' \emph{IEEE
  Transactions on Affective Computing}, vol.~5, no.~4, pp. 377--390, 2014.

\bibitem{dataset:MSP-IMPROV}
\BIBentryALTinterwordspacing
C.~Busso, S.~Parthasarathy, A.~Burmania, M.~AbdelWahab, N.~Sadoughi, and E.~M.
  Provost, ``{MSP}-{IMPROV}: An acted corpus of dyadic interactions to study
  emotion perception,'' \emph{{IEEE} Transactions on Affective Computing},
  vol.~8, no.~1, pp. 67--80, Jan. 2017. [Online]. Available:
  \url{https://doi.org/10.1109/taffc.2016.2515617}
\BIBentrySTDinterwordspacing

\bibitem{dataset:EmoV-DB}
\BIBentryALTinterwordspacing
A.~Adigwe, N.~Tits, K.~E. Haddad, S.~Ostadabbas, and T.~Dutoit, ``The emotional
  voices database: Towards controlling the emotion dimension in voice
  generation systems,'' \emph{CoRR}, vol. abs/1806.09514, 2018. [Online].
  Available: \url{http://arxiv.org/abs/1806.09514}
\BIBentrySTDinterwordspacing

\bibitem{dataset:ESD}
K.~Zhou, B.~Sisman, R.~Liu, and H.~Li, ``Seen and unseen emotional style
  transfer for voice conversion with a new emotional speech dataset,'' in
  \emph{ICASSP 2021-2021 IEEE International Conference on Acoustics, Speech and
  Signal Processing (ICASSP)}.\hskip 1em plus 0.5em minus 0.4em\relax IEEE,
  2021, pp. 920--924.

\bibitem{dataset:common_voice}
\BIBentryALTinterwordspacing
R.~Ardila, M.~Branson, K.~Davis, M.~Henretty, M.~Kohler, J.~Meyer, R.~Morais,
  L.~Saunders, F.~M. Tyers, and G.~Weber, ``Common voice: {A}
  massively-multilingual speech corpus,'' \emph{CoRR}, vol. abs/1912.06670,
  2019. [Online]. Available: \url{http://arxiv.org/abs/1912.06670}
\BIBentrySTDinterwordspacing

\bibitem{book:ekman_1999}
T.~Dalgleish and M.~J. Power, Eds., \emph{Handbook of cognition and
  emotion}.\hskip 1em plus 0.5em minus 0.4em\relax Chichester, England ; New
  York: Wiley, 1999.

\bibitem{paper:shaver_emotion_1987}
\BIBentryALTinterwordspacing
P.~Shaver, J.~Schwartz, D.~Kirson, and C.~O'Connor,
  ``\BIBforeignlanguage{en}{Emotion knowledge: {Further} exploration of a
  prototype approach.}'' \emph{\BIBforeignlanguage{en}{Journal of Personality
  and Social Psychology}}, vol.~52, no.~6, pp. 1061--1086, 1987. [Online].
  Available:
  \url{http://doi.apa.org/getdoi.cfm?doi=10.1037/0022-3514.52.6.1061}
\BIBentrySTDinterwordspacing

\bibitem{paper:MFA}
M.~McAuliffe, M.~Socolof, S.~Mihuc, M.~Wagner, and M.~Sonderegger, ``{Montreal
  Forced Aligner: Trainable Text-Speech Alignment Using Kaldi},'' in
  \emph{Proc. Interspeech 2017}, 2017, pp. 498--502.

\bibitem{license:apache_2004}
\BIBentryALTinterwordspacing
``Apache {License}, {Version} 2.0,'' 2004, publisher: Apache. [Online].
  Available: \url{https://www.apache.org/licenses/LICENSE-2.0}
\BIBentrySTDinterwordspacing

\end{thebibliography}

\end{document}